\documentclass{article}
\usepackage{spconf,amsmath,graphicx}
\usepackage{hyperref}
\usepackage{array}
\usepackage{multirow}
\usepackage{ragged2e}
\usepackage{tabularx}
\usepackage{newtxtext,newtxmath}
\usepackage[T1]{fontenc}
\usepackage{lmodern}

\usepackage{enumitem}
\setlist{nosep, leftmargin=14pt}

\usepackage{mwe} 

\title{Generalize Polyp Segmentation via Inpainting across Diverse Backgrounds and Pseudo-Mask Refinement}
\name{Jiajian Ma$^{1,2}$ \qquad Fangqi Lu$^{4}$ \qquad Silin Huang$^{4}$ \qquad Song Wu$^{4}$ \qquad Zhen Li$^{3,1}$\sthanks{Corresponding Author: Prof. Zhen Li, Email: lizhen@cuhk.edu.cn}}

\address{$^{1}$ FNII, CUHK-Shenzhen  \qquad
    $^{2}$ SDS, CUHK-Shenzhen \qquad
    $^{3}$ SSE, CUHK-Shenzhen \\
    $^{4}$  South China Hospital, Medical School, Shenzhen University  \\
    }

\begin{document}

%

\maketitle

\begin{abstract}

Inpainting lesions within different normal backgrounds is a potential method of addressing the generalization problem, which is crucial for polyp segmentation models. However, seamlessly introducing polyps into complex endoscopic environments while simultaneously generating accurate pseudo-masks remains a challenge for current inpainting methods. To address these issues, we first leverage the pre-trained Stable Diffusion Inpaint and ControlNet, to introduce a robust generative model capable of inpainting polyps across different backgrounds. Secondly, we utilize the prior that synthetic polyps are confined to the inpainted region, to establish an inpainted region-guided pseudo-mask refinement network. We also propose a sample selection strategy that prioritizes well-aligned and hard synthetic cases for further model fine-tuning. Experiments demonstrate that our inpainting model outperformed baseline methods both qualitatively and quantitatively in inpainting quality. Moreover, our data augmentation strategy significantly enhances the performance of polyp segmentation models on external datasets, achieving or surpassing the level of fully supervised training benchmarks in that domain. Our code is available at \href{https://github.com/497662892/PolypInpainter}{https://github.com/497662892/PolypInpainter}.

\end{abstract}
\section{Introduction}

Polyp detection and segmentation models can effectively improve polyp detection rates in endoscopic examinations \cite{gong2020detection}. However, constrained by data privacy and annotation costs, these models are predominantly trained on labeled data from limited sources  \cite{gong2020detection}. These datasets usually inadequately represent the diversity of real clinical settings, leading to a domain gap and a potential performance drop in application \cite{dong2021polyp}. 

\begin{figure}[htb]

\begin{minipage}[b]{.49\linewidth}
  \centering
  \centerline{\includegraphics[width=2.5cm]{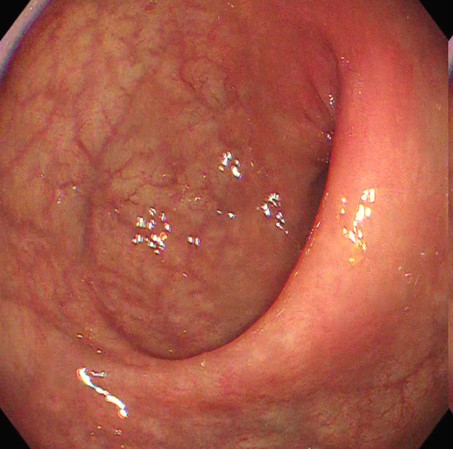}} 
  \vspace{-0.1cm} 
  \centerline{(a) External background}\medskip
\end{minipage}
\hfill
\begin{minipage}[b]{.49\linewidth}
  \centering
  \centerline{\includegraphics[width=2.5 cm]{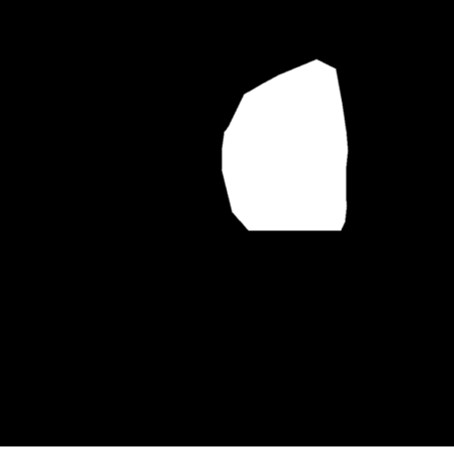}}
  \vspace{-0.1cm}
  \centerline{(b) Conditional mask}\medskip
\end{minipage}
\begin{minipage}[b]{.49\linewidth}
  \centering
  \centerline{\includegraphics[width=2.5 cm]{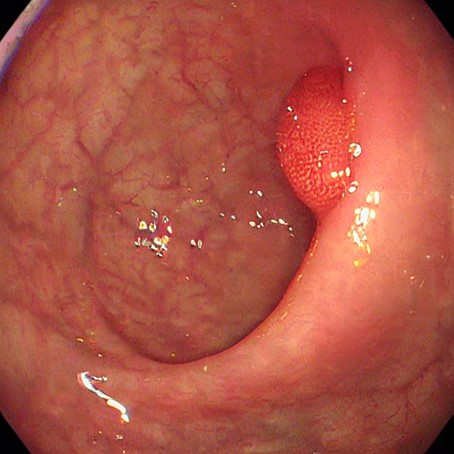}}
 \vspace{-0.1cm}
  \centerline{(c) Generated image}\medskip
\end{minipage}
\hfill
\begin{minipage}[b]{0.49\linewidth}
  \centering
  \centerline{\includegraphics[width=2.5 cm]{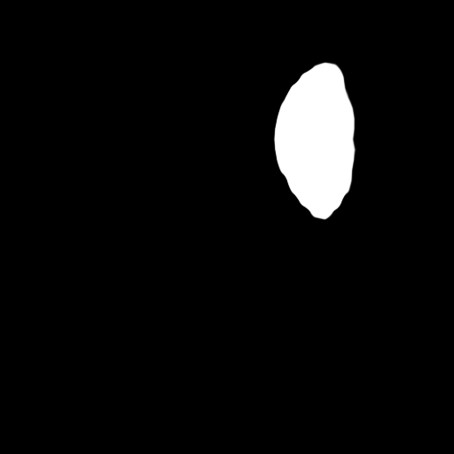}}
 \vspace{-0.1cm}
  \centerline{(d) Refined mask}\medskip
\end{minipage}
\vspace{-0.5cm}
\caption{Example of inpainting in external background and pseudo-mask refinement.}
\label{fig:1}
\vspace{-0.5cm}
\end{figure}

Compared with the limited number of labeled positive samples, negative images are much more abundant in clinical practice and face fewer privacy problems. Considering the polyp typically occupies less than 5\% of an endoscopic image \cite{dong2021polyp}, these negative images might contain rich information about the distribution of the external datasets. Thus, inpainting polyps into diverse backgrounds (Figure \ref{fig:1}) appears to be powerful in bridging the domain gaps.

Currently, many different content-introducing inpainting methods have been proposed \cite{perez2023poisson, zhou2023spatially, shen2023image, rombach2022high}, and some have been applied in data augmentation, including copy-paste \cite{zhou2023spatially} and GAN-based \cite{shen2023image} inpainting methods, and have been proven to increase model performance within internal datasets effectively. But surprisingly, their effectiveness on model generalization has not been thoroughly investigated.

Considering the ability to simulate the distribution of external datasets is key to data augmentation, diffusion-based inpainting models \cite{rombach2022high}, known for producing more realistic and diverse images than other inpainting methods, may excel in this area. However, their application in segmentation data augmentation is limited by the challenge of how to obtain high-quality pseudo-masks for synthetic images. 

To address these issues, we first propose the Polyp Inpainter, a diffusion-based polyp inpainting model with fine-grained boundary and surface features injected via ControlNet \cite{zhang2023adding} modules, which can realistically introduce polyps into diverse backgrounds. We also leveraged the prior that synthetic polyps are confined to the inpainted region and designed an inpainted region-guided segmentation network to generate high-quality pseudo-masks (Figure \ref{fig:1}, d). Additionally, we developed a sample selection strategy that prioritizes high-quality and challenging synthetic images for further segmentation model fine-tuning. Extensive experiments demonstrate that our method significantly improves model performance in external datasets, achieving or surpassing the level of benchmarks trained and validated with full supervision in that domain.

\section{METHODOLOGY}

\subsection{Diffusion-based polyp inpainting model}

\begin{figure}[htb]

\begin{minipage}[b]{0.95\linewidth}
  \centering
  \centerline{\includegraphics[width=8.5cm]{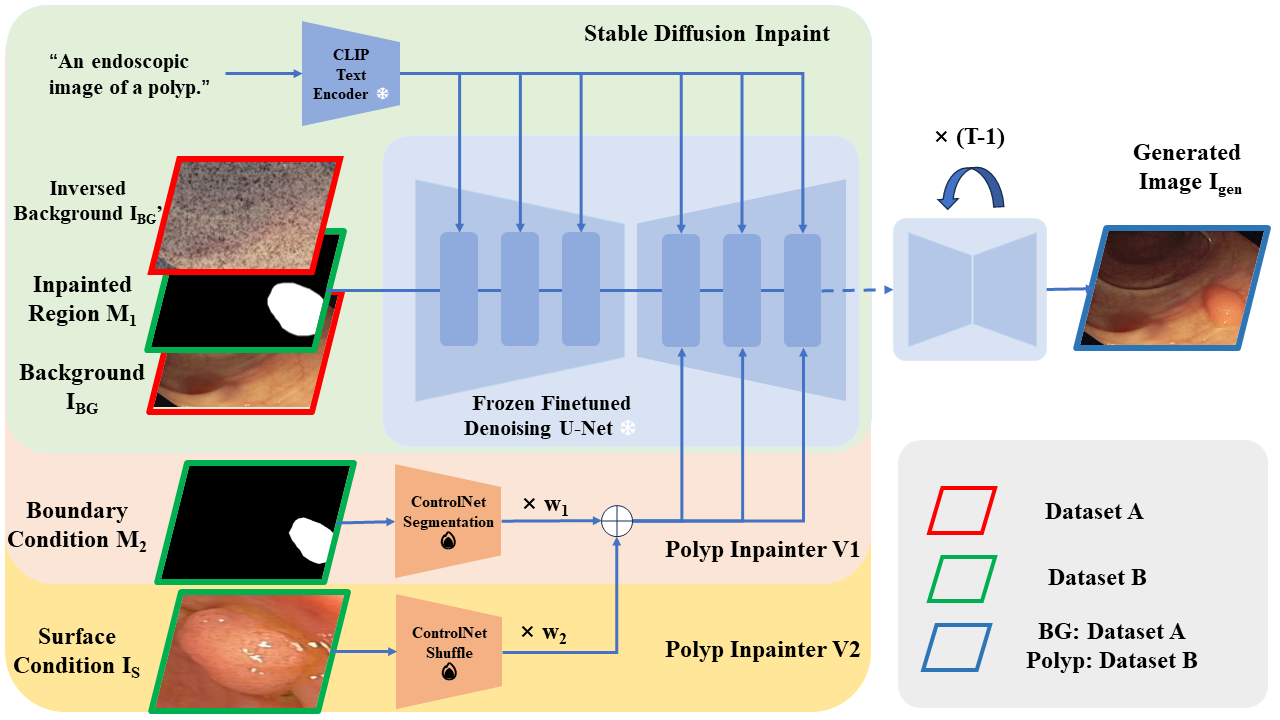}}
  \centerline{(a) The structure of our polyp inpainting model}\medskip
\end{minipage}

\begin{minipage}[b]{0.95\linewidth}
  \centering
  \centerline{\includegraphics[width=8.5cm]{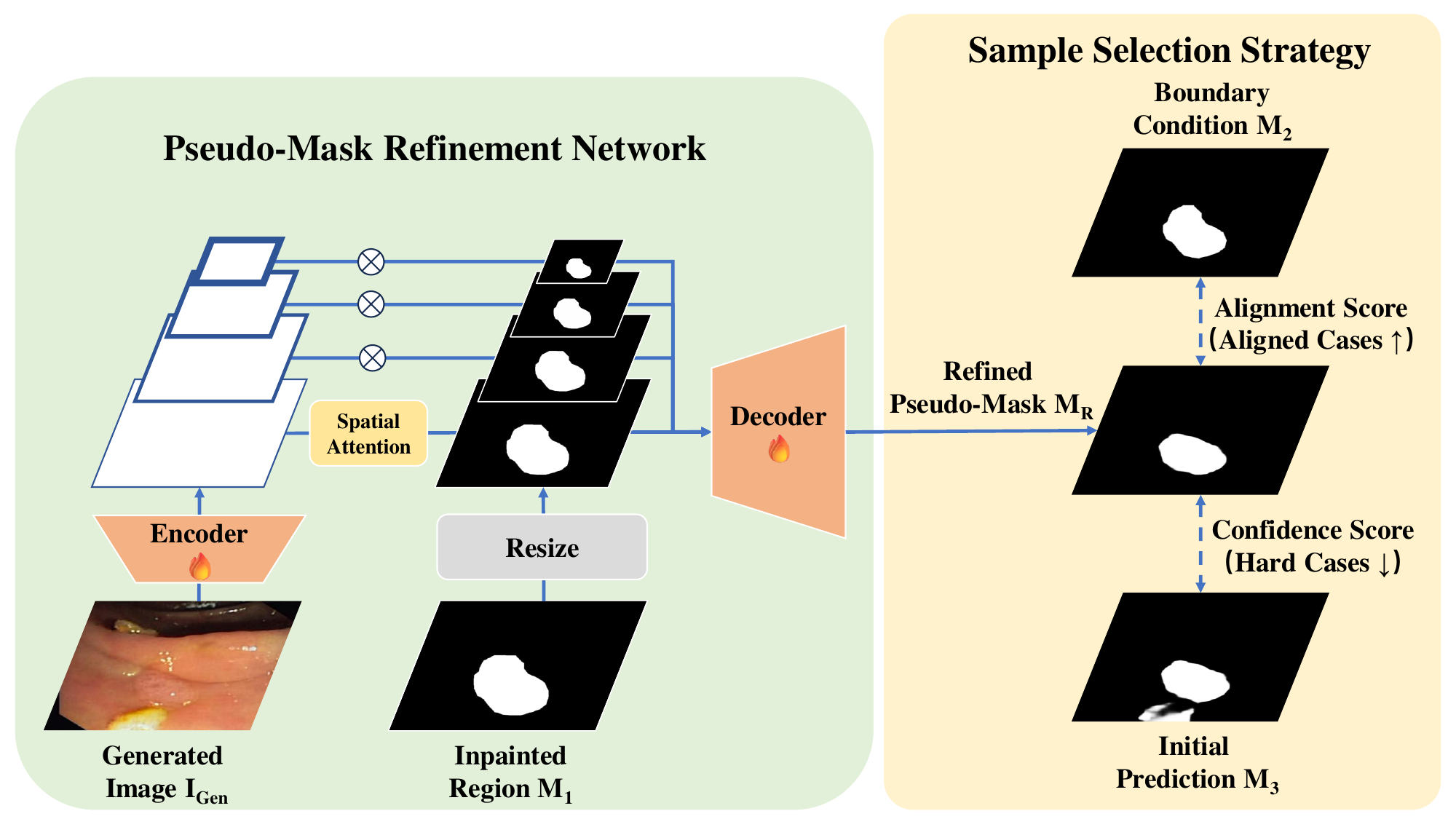}}
  \centerline{(b) Pseudo-mask refinement and case selection strategy}\medskip
\end{minipage}
 \vspace{-0.2cm}
\caption{The framework of our proposed data augmentation method.}
\label{fig:2}

 \vspace{-0.5cm}
\end{figure}

Figure \ref{fig:2}(a) displays the architecture of our proposed polyp inpainting model. We employ the Stable Diffusion Inpaint v1.5 as its backbone \cite{rombach2022high}, which will be frozen, after fine-tuning on an endoscopic dataset (Figure \ref{fig:2}(a), green). We further add and fine-tune 2 ControlNet modules, which are initialized by segmentation v1.1 or shuffle v1.1 \cite{zhang2023adding} to inject boundary or surface features, utilizing real mask \( M_2 \) or cropped polyp images \( I_S \) as conditions respectively. Based on whether injecting surface features, Polyp Inpainter is differentiated as V1 (Figure \ref{fig:2}(a), red) and V2 (Figure \ref{fig:2}(a), yellow). 

To enhance the naturalness of the generated images \( I_{Gen} \), we move the centroid of the inpainted region \( M_1 \) and boundary condition \( M_2 \) to the center of a patch from the background \( I_{BG}\) that closely resembles the polyp's color. Furthermore, we try to sample from an inversed noisy background \( I_{BG}' \) \cite{meng2021sdedit}as a superior alternative to the default standard Gaussian noise (Figure \ref{fig:2}(a), green). 

\subsection{Pseudo-mask refinement network}

The structure of the pseudo-mask refinement network is shown in Figure \ref{fig:2} (b, green block). In our generated images \( I_{Gen} \), polyps are exclusively present within the inpainted region \( M_1 \). Leveraging this fact, we gate the multi-scale encoder features with either element-wise multiplication or spatial attention with \( M_1 \). Consequently, it can filter out distractions from the complex endoscopic background, and produce high-quality refined pseudo-mask \( M_R \). 

\subsection{Sample selection strategy}

After obtaining the pseudo-mask \( M_R \), we further introduce a strategy for selecting high-quality and challenging samples, as depicted in Figure \ref{fig:2}(b, yellow). We define a metric, the alignment score \( S_{Align} \), as the Dice(\( M_R \), \( M_2 \)), to measure the alignment of the generated polyps with their boundary conditional \( M_2 \). Samples with \( S_{Align} \) \( \geq 0.93 \) are classified as well-aligned cases.

Concurrently, we define confidence score \( S_{Confident} \), as the Dice(\( M_R \), \( M_3 \)), to measure the difficulty of generated samples by comparing pseudo-mask \( M_R \) with the segmentation model's initial prediction \( M_3 \). Those with \( S_{Confident} \) \( \leq 0.9 \) are categorized as hard cases. This bifurcated approach allows us to selectively enrich our dataset with samples that are both well-aligned and challenging for further fine-tuning.

\section{EXPERIMENTS AND DISCUSSION}

\subsection{Datasets}

Our study used four public datasets for polyp segmentation: Kvasir-SEG \cite{jha2020kvasir} (1,000 images), ETIS-LaribPolypDB \cite{silva2014toward} (196 images), SUN-SEG \cite{misawa2021development, ji2022video} (49,136 images) and CVC-EndoSceneStill \cite{vazquez2017benchmark} (912 images). The first two were randomly split into 60:20:20 for training, validation, and testing. To have a quick evaluation, we randomly sampled SUN-SEG train/validation at 1/30 frame rate (848 and 186 images) and its official testing set at 1/10, leading to 2 testing subsets named SUN-SEG-Easy sub (1,742 images) and SUN-SEG-Hard sub (1,274 images). CVC-EndoSceneStill was divided as per existing guidelines \cite{vazquez2017benchmark}.

To serve as backgrounds for polyp inpainting, 761 polyp-free images from SUN-SEG were included. Since negative images were missing from the other datasets, we used a diffusion inpaint model to remove small polyps from some train/validation images, adding 64, 154, and 74 negative images to Kvasir-SEG, CVC-EndoSceneStill, and ETIS-LaribPolypDB, respectively.

\begin{table}[ht]
\centering
\caption{Training parameters for inpainting models}
\begin{tabularx}{\columnwidth}{lXXX}
\hline
\textbf{Model} & \textbf{Steps} & \textbf{LR} & \textbf{Batch Size} \\ \hline
SD Inpaint & 2,000 & \(1 \times 10^{-5}\) & 4 \\
V1 ControlNet & 15,000 & \(5 \times 10^{-5}\) & 4 \\
V2 ControlNet & 16,000 & \(5 \times 10^{-5}\) & 2 \\ \hline
\end{tabularx}
\label{table:training_params}
\vspace{-0.5cm}
\end{table}

\subsection{Implementation details}

We utilized a single RTX 4090 GPU and the PyTorch framework for all experiments. Our inpainting models were exclusively fine-tuned on the Kvasir-SEG train/val set with the default AdamW, using hyperparameters listed in table \ref{table:training_params}. During inference, we sampled from a noisy background at an inverse strength of 0.85 \cite{meng2021sdedit} and employed the UniPCMultistepScheduler sampler \cite{zhao2023unipc} to generate the inpainting images by 50 sampling steps. 

We adapted the polyp-PVT model, a robust baseline \cite{dong2021polyp}, into a pseudo-mask refinement network for our study. This model was then trained on the Kvasir-SEG dataset, adhering to the protocols outlined by Dong et al. \cite{dong2021polyp}. The model demonstrating the highest performance during the validation was used to refine the pseudo-masks of synthetic images.

In our experiment, we utilized masks (w/wo cropped images) from the Kvasir-SEG train/val sets as conditions for inpainting, integrating them with 10 distinct negative images from each of the 4 different datasets. This process resulted in the generation of 32,000 synthetic images. Following the implementation of our sample selection strategy, 100-300 cases with background from each dataset were selected, consisting of our well-aligned and hard synthetic dataset. 

The training of the polyp-PVT segmentation model in our pipeline is executed in two distinct stages. Initially, we adhere to the established protocol \cite{dong2021polyp} to train on the Kvasir-SEG dataset. This is followed by a phase of fine-tuning on a merged dataset, comprising both the Kvasir-SEG and the synthetic data. The fine-tuning process extends up to 25 epochs, employing a learning rate of \(1 \times 10^{-5}\). The best model in validation was then selected for testing.

\subsection{Evaluation of polyp inpainting quality}

\begin{figure}[htb]

\begin{minipage}[b]{.13\linewidth}
  \centering
  \centerline{BG}\medskip
  \centerline{\includegraphics[width=1.4cm]{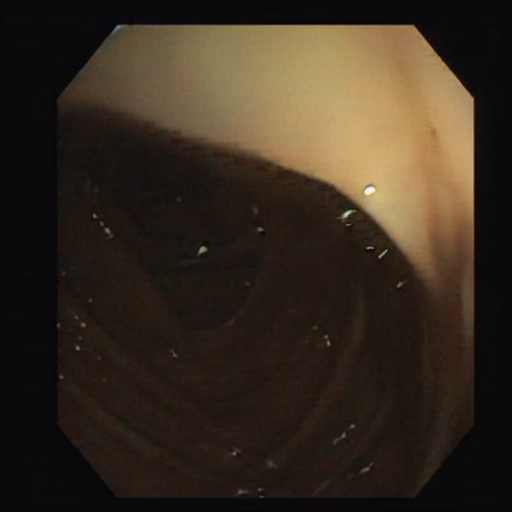}}
\end{minipage}
\hfill
\begin{minipage}[b]{.13\linewidth}
  \centering
  \centerline{Ref}\medskip
  \centerline{\includegraphics[width=1.4cm]{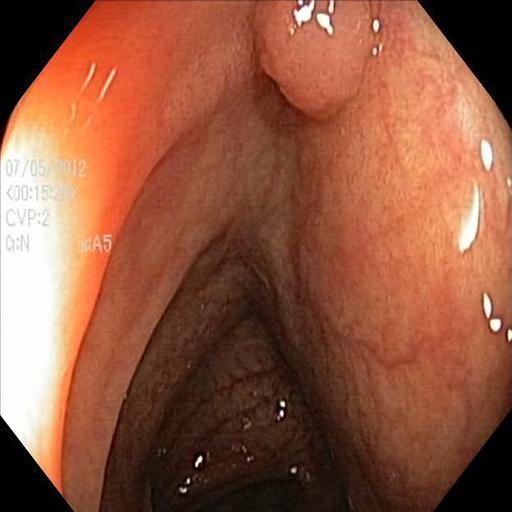}}
\end{minipage}
\hfill
\begin{minipage}[b]{.13\linewidth}
  \centering
  \centerline{Ours V2}\medskip
  \centerline{\includegraphics[width=1.4cm]{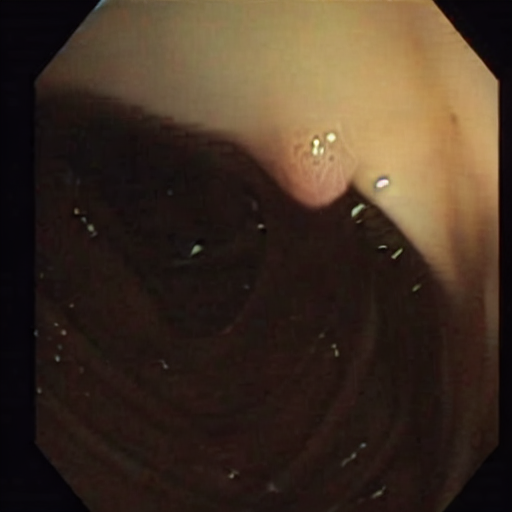}}
\end{minipage}
\hfill
\begin{minipage}[b]{.13\linewidth}
  \centering
  \centerline{Ours V1}\medskip
  \centerline{\includegraphics[width=1.4cm]{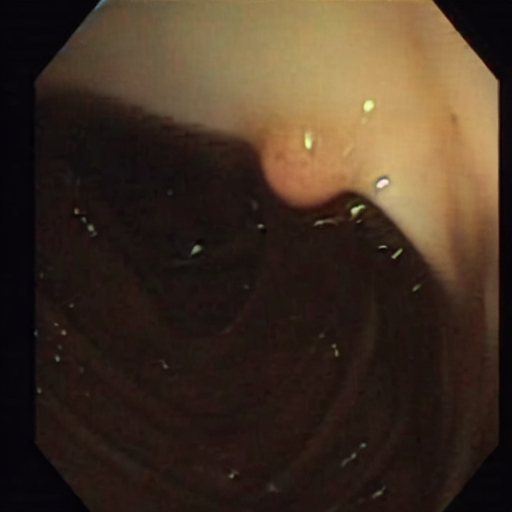}}
\end{minipage}
\hfill
\begin{minipage}[b]{.13\linewidth}
  \centering
  \centerline{SD}\medskip
  \centerline{\includegraphics[width=1.4cm]{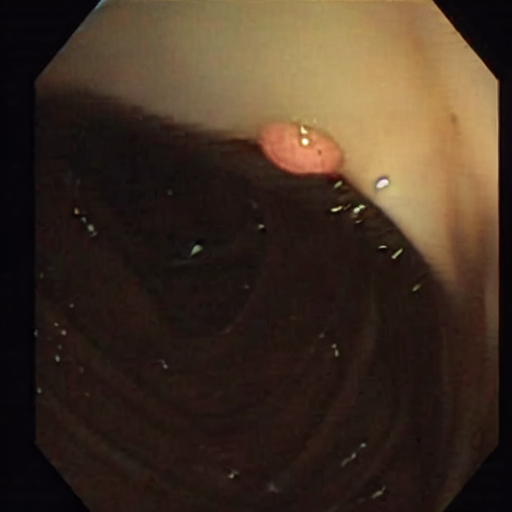}}
\end{minipage}
\hfill
\begin{minipage}[b]{.13\linewidth}
  \centering
  \centerline{Poisson}\medskip
  \centerline{\includegraphics[width=1.4cm]{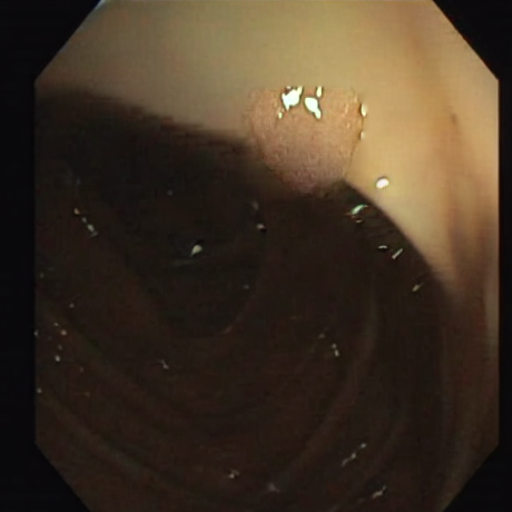}}
\end{minipage}

\begin{minipage}[b]{.13\linewidth}
  \centering
  \centerline{\includegraphics[width=1.4cm]{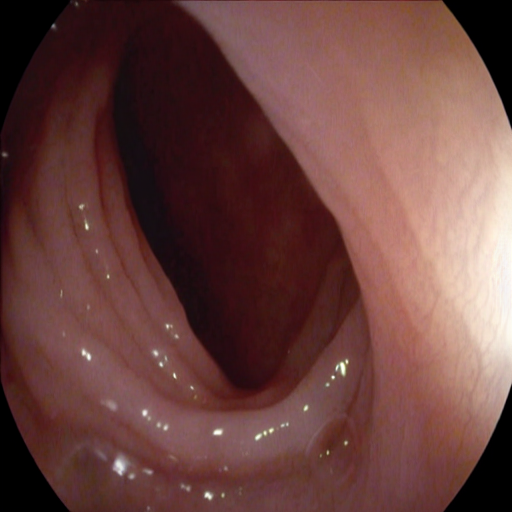}}
\end{minipage}
\hfill
\begin{minipage}[b]{.13\linewidth}
  \centering
  \centerline{\includegraphics[width=1.4cm]{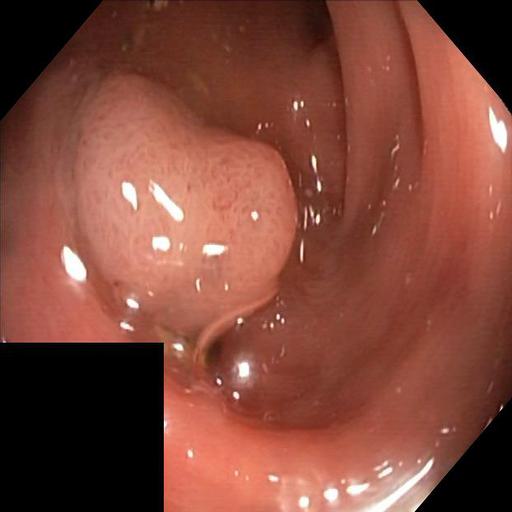}}
\end{minipage}
\hfill
\begin{minipage}[b]{.13\linewidth}
  \centering
  \centerline{\includegraphics[width=1.4cm]{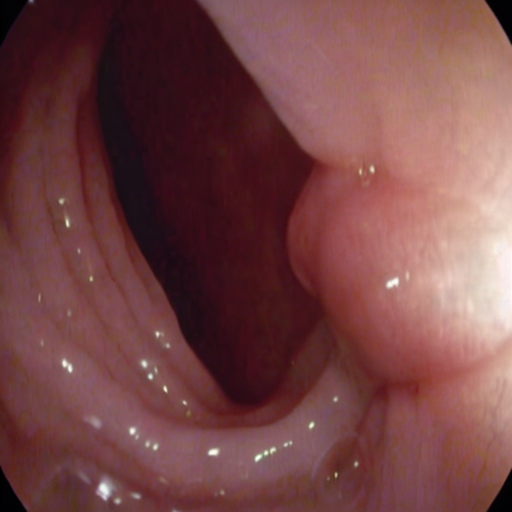}}
\end{minipage}
\hfill
\begin{minipage}[b]{.13\linewidth}
  \centering
  \centerline{\includegraphics[width=1.4cm]{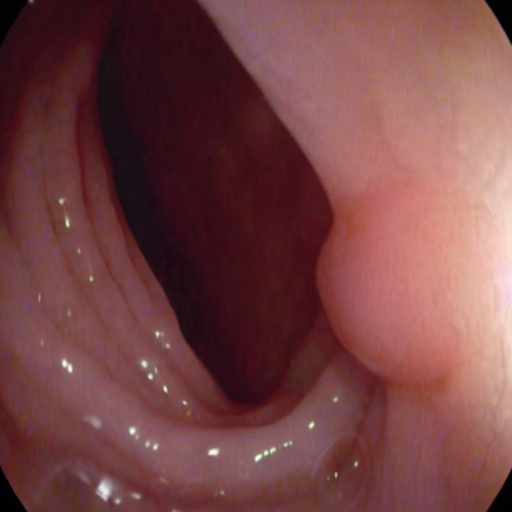}}
\end{minipage}
\hfill
\begin{minipage}[b]{.13\linewidth}
  \centering
  \centerline{\includegraphics[width=1.4cm]{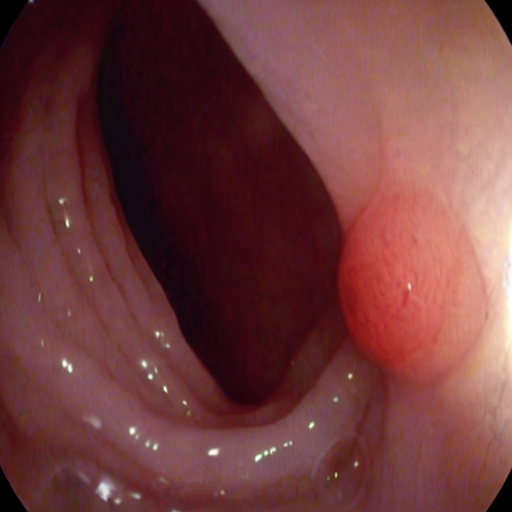}}
\end{minipage}
\hfill
\begin{minipage}[b]{.13\linewidth}
  \centering
  \centerline{\includegraphics[width=1.4cm]{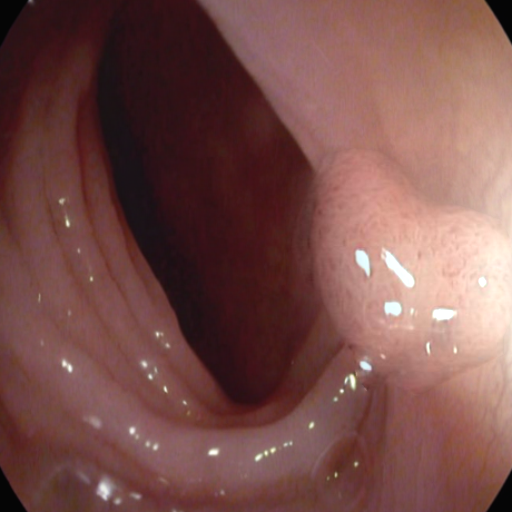}}
\end{minipage}

\begin{minipage}[b]{.13\linewidth}
  \centering
  \centerline{\includegraphics[width=1.4cm]{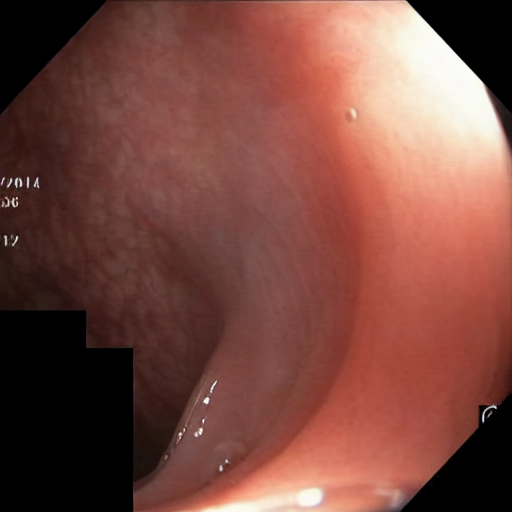}}
\end{minipage}
\hfill
\begin{minipage}[b]{.13\linewidth}
  \centering
  \centerline{\includegraphics[width=1.4cm]{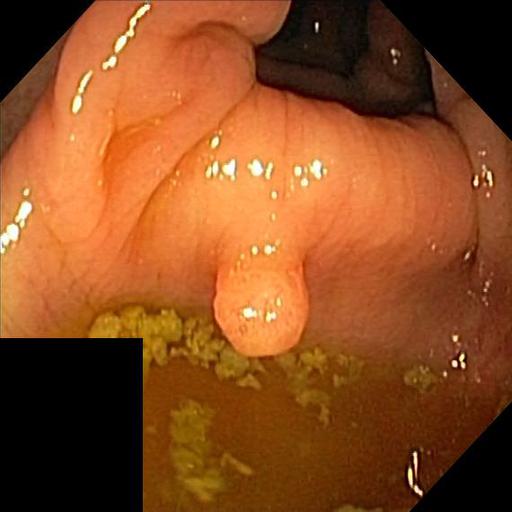}}
\end{minipage}
\hfill
\begin{minipage}[b]{.13\linewidth}
  \centering
  \centerline{\includegraphics[width=1.4cm]{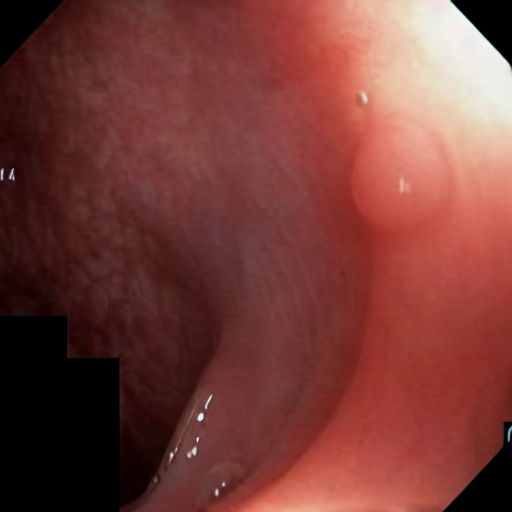}}
\end{minipage}
\hfill
\begin{minipage}[b]{.13\linewidth}
  \centering
  \centerline{\includegraphics[width=1.4cm]{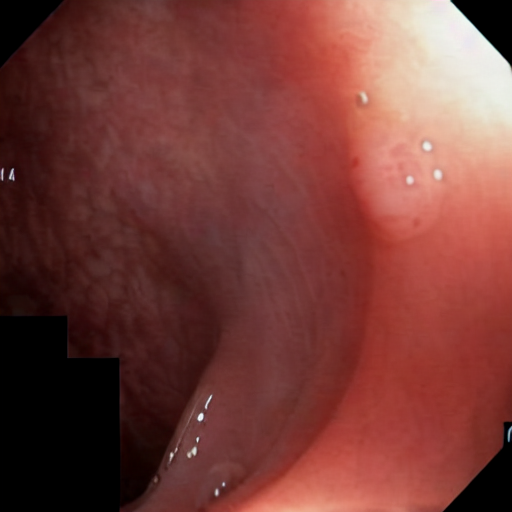}}
\end{minipage}
\hfill
\begin{minipage}[b]{.13\linewidth}
  \centering
  \centerline{\includegraphics[width=1.4cm]{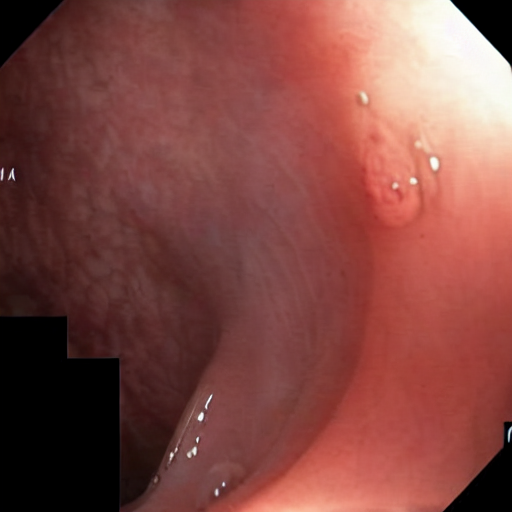}}
\end{minipage}
\hfill
\begin{minipage}[b]{.13\linewidth}
  \centering
  \centerline{\includegraphics[width=1.4cm]{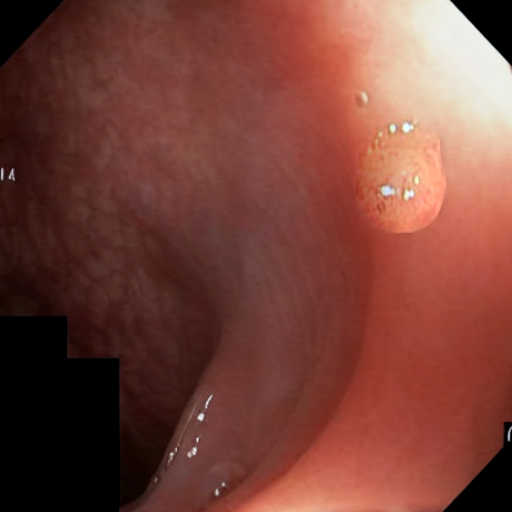}}
\end{minipage}

\begin{minipage}[b]{.13\linewidth}
  \centering
  \centerline{\includegraphics[width=1.4cm]{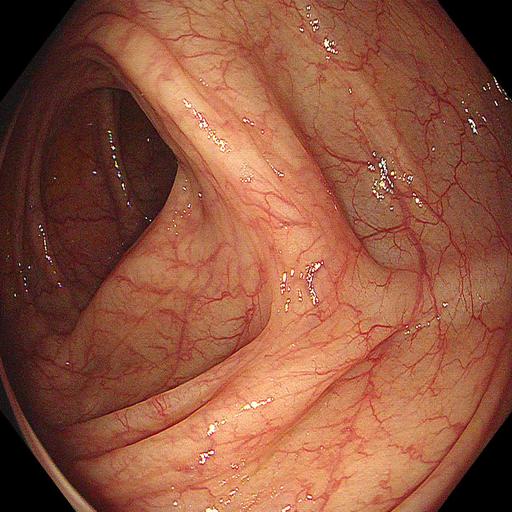}}
\end{minipage}
\hfill
\begin{minipage}[b]{.13\linewidth}
  \centering
  \centerline{\includegraphics[width=1.4cm]{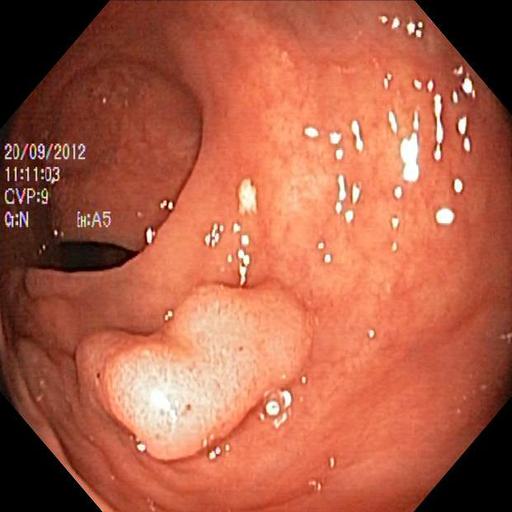}}
\end{minipage}
\hfill
\begin{minipage}[b]{.13\linewidth}
  \centering
  \centerline{\includegraphics[width=1.4cm]{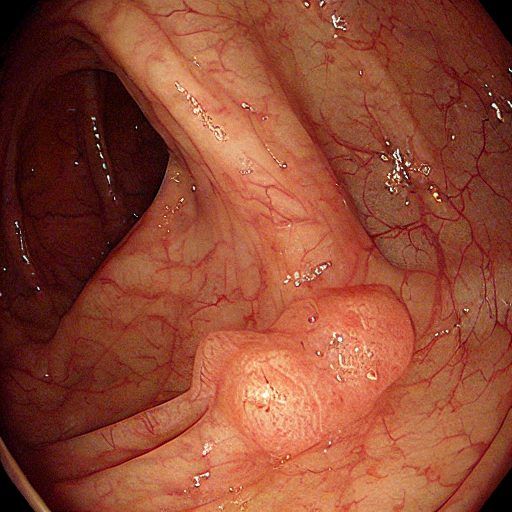}}
\end{minipage}
\hfill
\begin{minipage}[b]{.13\linewidth}
  \centering
  \centerline{\includegraphics[width=1.4cm]{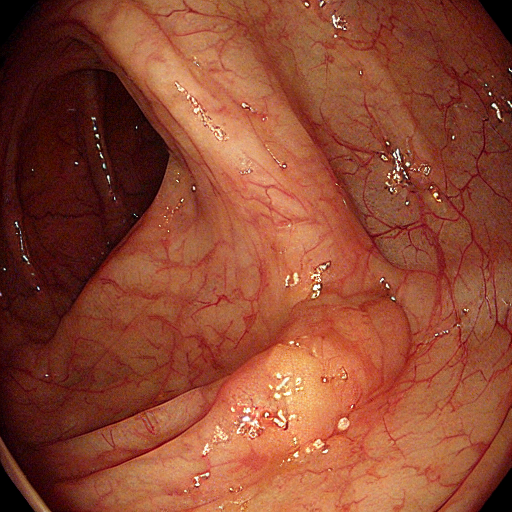}}
\end{minipage}
\hfill
\begin{minipage}[b]{.13\linewidth}
  \centering
  \centerline{\includegraphics[width=1.4cm]{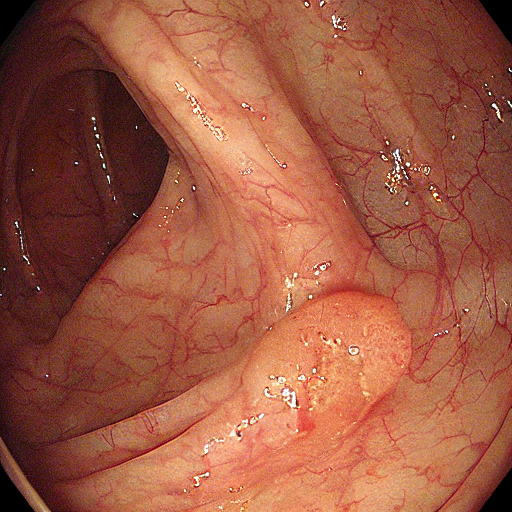}}
\end{minipage}
\hfill
\begin{minipage}[b]{.13\linewidth}
  \centering
  \centerline{\includegraphics[width=1.4cm]{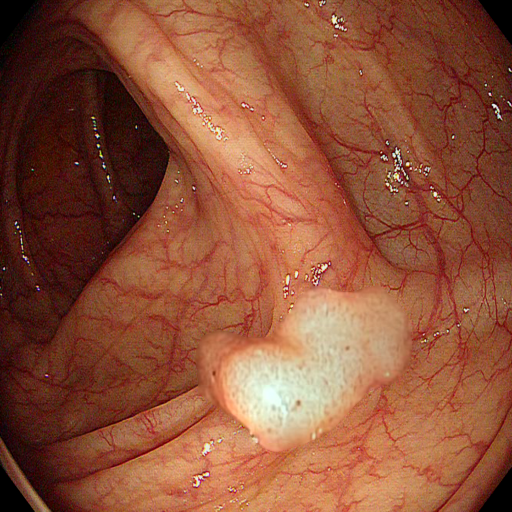}}
\end{minipage}
\caption{Qualitative results of polyp inpainting methods. BG: background, Ref: reference, SD: Stable Diffusion Inpaint, Poisson: Poisson image blending.}
\label{fig:3}
\vspace{-0.5cm}
\end{figure}

Figure \ref{fig:3} presents a qualitative comparison among different inpainting methods, demonstrating the diffusion-based methods' superiority in generating realistic polyps (col 3, 4, 5). Nevertheless, our Polyp Inpainter V1 and V2, exhibit more natural results, with better boundary harmony, mask alignment, and higher reference similarity than the baseline Stable Diffusion Inpaint model. Further examination of V1 and V2 shows both versions achieve comparable levels of naturalness, but V2 is more adept at capturing surface features and maintaining similarity to the reference images (col 3, 4).

For quantitative evaluation, two medical experts were invited to evaluate the naturalness and similarity of 120 sets of synthesized images independently. Each set consisted of 5 images generated with the same background and inpainted region using different methods: copy-paste, Poisson image blending \cite{perez2023poisson}, Stable Diffusion Inpaint, and Polyp Inpainter V1 and V2. As indicated in Table \ref{table:1}, our models significantly outperform the others, with V2 obtaining the best average rankings in both naturalness (1.562) and similarity (1.479).

Additionally, average alignment scores calculated from 32,000 generated images further confirm our models' superior capability of boundary control, with V1 scoring 0.847 and V2 scoring 0.864, which are much higher than the baseline Stable Diffusion Inpaint model (0.754).

\begin{table}

\footnotesize
\setlength{\tabcolsep}{1.5pt} 
\renewcommand{\arraystretch}{1.2} 

\centering
\caption{Quantitative evaluation on inpainting image quality.}
\begin{tabular}{>{\hspace{0pt}}m{0.269\linewidth}|>{\centering\hspace{0pt}}m{0.219\linewidth}>{\centering\hspace{0pt}}m{0.233\linewidth}>{\centering\arraybackslash\hspace{0pt}}m{0.2\linewidth}} 
\hline
                       & Avg Naturalness Ranking ↓ & Avg Similarity Ranking ↓ & Alignment Score ↑  \\ 
\hline
Copy Paste             &         4.988          & -                    & -                  \\
Poisson Blending &            3.329         & -                    & -                  \\
SD Inpaint             &         2.204            &     2.354                 & 0.754              \\
Ours V1                &          1.775           &      1.558                & 0.847              \\
Ours V2                &         \textbf{1.562}           &  \textbf{1.479}                    & \textbf{0.864}              \\
\hline
\end{tabular}
\label{table:1}
\vspace{-0.5cm}
\end{table}

\subsection{Generalization in polyp segmentation }

Table \ref{table:2} displays the polyp-PVT's performance across diverse datasets and augmentation strategies. The models trained on the Kvasir-SEG show a performance reduction on external datasets (row 1) when compared with the performance of benchmarks training and validating in that domain (see row 4). This drop is pronounced in the ETIS-LaribPolypDB (mDice: 0.729 v.s. 0.774), SUN-SEG-Easy sub (mDice: 0.806 v.s. 0.827), and SUN-SEG-Hard sub (mDice: 0.789 v.s. 0.817) datasets, highlighting a substantial generalization problem for models trained solely on Kvasir-SEG.

\begin{table*}

\footnotesize

\setlength{\tabcolsep}{1.5pt} 
\renewcommand{\arraystretch}{1.2} 

\centering
\vspace{-0.2cm}
\caption{The performance of polyp-PVT across different datasets and augmentation methods}
\begin{tabular}{>{\hspace{0pt}}m{0.078\linewidth}>{\hspace{0pt}}m{0.078\linewidth}|>{\centering\hspace{0pt}}m{0.0625\linewidth}>{\centering\hspace{0pt}}m{0.0625\linewidth}>{\centering\hspace{0pt}}m{0.0625\linewidth}>{\centering\hspace{0pt}}m{0.0625\linewidth}>{\centering\hspace{0pt}}m{0.0625\linewidth}>{\centering\hspace{0pt}}m{0.0625\linewidth}>{\centering\hspace{0pt}}m{0.0625\linewidth}>{\centering\hspace{0pt}}m{0.0625\linewidth}>{\centering\hspace{0pt}}m{0.0625\linewidth}>{\centering\hspace{0pt}}m{0.0625\linewidth}|>{\centering\hspace{0pt}}m{0.046\linewidth}>{\centering\arraybackslash\hspace{0pt}}m{0.046\linewidth}} 
\hline
\multicolumn{1}{>{\hspace{0pt}}m{0.078\linewidth}}{\multirow{2}{0.078\linewidth}{Train Dataset}} & \multicolumn{1}{>{\hspace{0pt}}m{0.078\linewidth}|}{\multirow{2}{0.078\linewidth}{Aug}} & \multicolumn{2}{>{\centering\hspace{0pt}}m{0.125\linewidth}}{EndoSceneStill~} & \multicolumn{2}{>{\centering\hspace{0pt}}m{0.125\linewidth}}{LaribPolypDB} & \multicolumn{2}{>{\centering\hspace{0pt}}m{0.125\linewidth}}{Kvasir-SEG} & \multicolumn{2}{>{\centering\hspace{0pt}}m{0.125\linewidth}}{SUN-SEG Easy Sub} & \multicolumn{2}{>{\centering\hspace{0pt}}m{0.125\linewidth}|}{SUN-SEG Hard Sub} & \multicolumn{2}{>{\centering\arraybackslash\hspace{0pt}}m{0.092\linewidth}}{Overall}  \\ 
\cline{3-14}
\multicolumn{1}{>{\hspace{0pt}}m{0.04\linewidth}}{}                                                         & \multicolumn{1}{>{\hspace{0pt}}m{0.106\linewidth}|}{}                                                        & mDice                  & mIoU                                                & mDice                  & mIoU                                             & mDice                  & mIoU                                       & mDice                  & mIoU                                                & mDice                  & mIoU                                                  & mDice          & mIoU                                                                 \\ 
\hline
Kvasir-SEG                                                                                                       & /                                                                                                            & 0.861 (0.006)          & 0.794 (0.005)                                       & 0.729 (0.042)          & 0.636 (0.043)                                    & 0.908 (0.007)          & 0.854 (0.010)                              & 0.806 (0.007)          & 0.737 (0.008)                                       & 0.789 (0.010)          & 0.715 (0.011)                                         & 0.819          & 0.747                                                                \\

Kvasir-SEG & Copy Paste & 0.861 (0.007)          & 0.791 (0.006) & 0.690 (0.040)          & 0.608 (0.041)  & \textbf{0.915 (0.003)} & \textbf{0.863 (0.002)}  & 0.753 (0.014)          & 0.685 (0.013)    & 0.733 (0.012)          & 0.661 (0.009)  & 0.790   & 0.722       \\
Kvasir-SEG  & SD Inpaint   & 0.870 (0.003)  & 0.803 (0.004)  & 0.759 (0.027)          & 0.676 (0.033)  & 0.911 (0.003) & 0.858 (0.003)    & 0.826 (0.014)  & 0.757 (0.015) & 0.797 (0.017)  & 0.721 (0.018)   & 0.833   & 0.763                                        \\ 

\hline

In Domain  & /  & 0.864 (0.011)   & 0.796 (0.015)  & 0.774 (0.047)  & 0.686 (0.048) & 0.908 (0.007) & 0.854 (0.010) & 0.827 (0.008)       & 0.756 (0.008) & \textbf{0.817 (0.010)}  & \textbf{0.740 (0.009)}    & 0.838   & 0.766                                                                \\

\hline
Kvasir-SEG                                                                                                       & Ours V1                                                                                & 0.870 (0.006)          & 0.805 (0.006)   & 0.783 (0.040)    & 0.697 (0.042)   & 0.914 (0.001)          & 0.863 (0.002)    & \textbf{0.838 (0.005)} & \textbf{0.771 (0.006)}   & 0.809 (0.005) & 0.736 (0.005)    & 0.843          & 0.774                         \\
Kvasir-SEG   & Ours V2  & \textbf{0.875 (0.006)} & \textbf{0.810 (0.006)}  & \textbf{0.810 (0.036)} & \textbf{0.727 (0.035)}   & 0.914 (0.004)          & 0.862 (0.005)  & 0.828 (0.009)          & 0.761 (0.010)      & 0.803 (0.008)          & 0.730 (0.008)                                         & \textbf{0.846} & \textbf{0.778}                                                       \\
\hline
\end{tabular}
\label{table:2}
\raggedright
\footnotesize{The numbers in parentheses represent the standard deviation of 5 experiments.}
\vspace{-0.5cm}
\end{table*}

Implementing our data augmentation method results in significant performance enhancements on both the Kvasir-SEG and external datasets. As indicated in row 5 of Table \ref{table:2}, augmentation with Polyp Inpainter V1 improved the mDice by 0.6\% on Kvasir-SEG (0.914 v.s. 0.908) and exhibited increments of 0.9\%, 5.4\%, 3.2\%, and 2.0\% on the CVC-EndoSceneStill, ETIS-LaribPolypDB, SUN-SEG-Easy sub, and SUN-SEG-Hard sub, respectively. Similarly, as shown in row 6, augmentation with Polyp Inpainter V2 increased the mDice by 0.6\% on Kvasir-SEG (0.914 v.s. 0.908) and led to improvements of 1.4\%, 8.1\%, 2.2\%, and 1.4\% on the corresponding external test sets. Both versions surpassed the in-domain training and validating models across all datasets, except for the SUN-SEG-Hard sub, where the results were relatively close. Furthermore, our augmentation method outperformed in-domain benchmarks and other augmentation techniques in the average performance of all datasets, with enhancements in overall average mDice of +0.5\% and +0.8\%, and mIoU of +0.8\% and +1.2\% for V1 and V2, respectively (last 2 cols). These findings affirm that our method can considerably boost generalization capabilities, by only using easily accessible background data from external datasets.

An additional noteworthy finding is that while the copy-paste approach may boost segmentation within the Kvasir-SEG test set, it significantly hampers generalization on external datasets (refer to row 2 of Table \ref{table:2}). The mDice scores for ETIS-LaribPolypDB, SUN-SEG-Easy sub, and SUN-SEG-Hard sub dropped by 3.9\%, 5.3\%, and 5.6\%, respectively. We conjecture that this reduction stems from the copy-paste method's failure to simulate lesions in the target domain's distribution, causing overfitting on the Kvasir-SEG dataset. This result implies that the quality of synthetic images makes sense in enhancing model generalization.

\subsection{Ablation study}

We conducted an ablation study to assess the individual contributions of each component in our data augmentation framework.  All images are generated by Polyp Inpainter V2, with an augmentation cap of 200 images per dataset for fair comparisons. Table \ref{table:3} (line 2) shows that utilizing the conditional mask \(M_2\) as a pseudo-mask directly improves model performance by +1.2\% in overall mDice and +0.9\% in overall mIoU, demonstrating the high quality of our inpainted images. Introducing pseudo-mask refinement (line 3) further enhances the segmentation model's performance by an additional +0.6\% in mDice and +1.2\% in mIoU, underscoring the refined pseudo-mask \(M_R\)'s advantage over \(M_2\) for segmentation tasks (see Figure \ref{fig:1}, b and d). Eliminating poorly aligned cases yields another increase of +0.6\% in mDice and +0.7\% in mIoU (line 4). By further selecting well-aligned yet challenging cases, we constructed our complete data augmentation framework (line 5), which achieved optimal performance, with an incremental rise of +0.3\% in both mDice and mIoU.

\begin{table}

\footnotesize
\setlength{\tabcolsep}{1.5pt} 
\renewcommand{\arraystretch}{1.2} 

\centering
\vspace{-0.2cm}
\caption{The ablation study on pseudo-mask refinement and sample selection strategy}
\begin{tabular}{>{\hspace{0pt}}m{0.083\linewidth}>{\hspace{0pt}}m{0.119\linewidth}>{\hspace{0pt}}m{0.119\linewidth}>{\hspace{0pt}}m{0.096\linewidth}|>{\hspace{0pt}}m{0.25\linewidth}>{\hspace{0pt}}m{0.25\linewidth}} 
\hline
Aug        & Refine     & Aligned      & Hard       & Overall mDice           & Overall mIoU             \\ 
\hline
-          & -          & -          & -          & 0.819                   & 0.747                    \\
+          & -          & -          & -          & 0.831 (+1.2\%)          & 0.756 (+0.9\%)           \\
+          & +          & -          & -          & 0.837 (+0.6\%)          & 0.768 (+1.2\%)           \\
+          & +          & +          & -          & 0.843 (+0.6\%)          & 0.775 (+0.7\%)           \\ 
\textbf{+} & \textbf{+} & \textbf{+} & \textbf{+} & \textbf{0.846 (+0.3\%)} & \textbf{0.778 (+0.3\%)}  \\
\hline
\end{tabular}
\label{table:3}
\vspace{-0.5cm}
\end{table}

\section{CONCLUSION}

Our study proposed an inpainting-based data augmentation strategy aimed at improving the generalization of polyp segmentation models, by introducing polyps into the normal background from different datasets. Qualitative and quantitative evaluations demonstrated our method can realistically inpaint polyps across various backgrounds. Extensive testing confirms that our approach consistently enhances model performance on external datasets, achieving or surpassing fully supervised training benchmarks in that domain. Considering the easy accessibility of negative endoscopy images, our method can potentially address the generalization challenges faced by polyp segmentation in real clinical settings.

\section{COMPLIANCE WITH ETHICAL STANDARDS}

This research study was conducted retrospectively using human subject data made available in open access by Simula Research Laboratory, Computer Vision Center, Showa University Northern Yokohama Hospital, and Mori Lab. Ethical approval was not required as confirmed by the license attached with the open-access data.

\section{Acknowledgments}

This work was supported in part by Shenzhen General Program No.JCYJ20220530143600001, by the Basic Research Project No.HZQB-KCZY-2021067 of Hetao Shenzhen HK S\&T Cooperation Zone, by Shenzhen-Hong Kong Joint Funding No.SGDX2021112311040202, by Shenzhen Outstanding Talents Training Fund, by Guangdong Research Project No.2017ZT07X152 and No.2019CX01X104, by the Guangdong Provincial Key Laboratory of Future Networks of Intelligence (Grant No.2022B1212010001), The Chinese University of Hong Kong, Shenzhen, by the NSFC 61931024\&81922046, by Tencent Open Fund.



\bibliographystyle{IEEEbib}
\bibliography{Template_ISBI_latex}

\begin{thebibliography}{10}

\bibitem{gong2020detection}
Dexin Gong, Lianlian Wu, Jun Zhang, Ganggang Mu, Lei Shen, Jun Liu, Zhengqiang Wang, Wei Zhou, Ping An, Xu~Huang, et~al.,
\newblock ``Detection of colorectal adenomas with a real-time computer-aided system (endoangel): a randomised controlled study,''
\newblock {\em The lancet Gastroenterology \& hepatology}, vol. 5, no. 4, pp. 352--361, 2020.

\bibitem{dong2021polyp}
Bo~Dong, Wenhai Wang, Deng-Ping Fan, Jinpeng Li, Huazhu Fu, and Ling Shao,
\newblock ``Polyp-pvt: Polyp segmentation with pyramid vision transformers,''
\newblock {\em arXiv preprint arXiv:2108.06932}, 2021.

\bibitem{perez2023poisson}
Patrick P{\'e}rez, Michel Gangnet, and Andrew Blake,
\newblock ``Poisson image editing,''
\newblock in {\em Seminal Graphics Papers: Pushing the Boundaries, Volume 2}, pp. 577--582. 2023.

\bibitem{zhou2023spatially}
Lei Zhou,
\newblock ``Spatially exclusive pasting: A general data augmentation for the polyp segmentation,''
\newblock in {\em 2023 International Joint Conference on Neural Networks (IJCNN)}. IEEE, 2023, pp. 01--07.

\bibitem{shen2023image}
Zhenrong Shen, Xi~Ouyang, Bin Xiao, Jie-Zhi Cheng, Dinggang Shen, and Qian Wang,
\newblock ``Image synthesis with disentangled attributes for chest x-ray nodule augmentation and detection,''
\newblock {\em Medical Image Analysis}, vol. 84, pp. 102708, 2023.

\bibitem{rombach2022high}
Robin Rombach, Andreas Blattmann, Dominik Lorenz, Patrick Esser, and Bj{\"o}rn Ommer,
\newblock ``High-resolution image synthesis with latent diffusion models,''
\newblock in {\em Proceedings of the IEEE/CVF conference on computer vision and pattern recognition}, 2022, pp. 10684--10695.

\bibitem{zhang2023adding}
Lvmin Zhang, Anyi Rao, and Maneesh Agrawala,
\newblock ``Adding conditional control to text-to-image diffusion models,''
\newblock in {\em Proceedings of the IEEE/CVF International Conference on Computer Vision}, 2023, pp. 3836--3847.

\bibitem{meng2021sdedit}
Chenlin Meng, Yutong He, Yang Song, Jiaming Song, Jiajun Wu, Jun-Yan Zhu, and Stefano Ermon,
\newblock ``Sdedit: Guided image synthesis and editing with stochastic differential equations,''
\newblock {\em arXiv preprint arXiv:2108.01073}, 2021.

\bibitem{jha2020kvasir}
Debesh Jha, Pia~H Smedsrud, Michael~A Riegler, P{\aa}l Halvorsen, Thomas de~Lange, Dag Johansen, and H{\aa}vard~D Johansen,
\newblock ``Kvasir-seg: A segmented polyp dataset,''
\newblock in {\em MultiMedia Modeling: 26th International Conference, MMM 2020, Daejeon, South Korea, January 5--8, 2020, Proceedings, Part II 26}. Springer, 2020, pp. 451--462.

\bibitem{silva2014toward}
Juan Silva, Aymeric Histace, Olivier Romain, Xavier Dray, and Bertrand Granado,
\newblock ``Toward embedded detection of polyps in wce images for early diagnosis of colorectal cancer,''
\newblock {\em International journal of computer assisted radiology and surgery}, vol. 9, pp. 283--293, 2014.

\bibitem{misawa2021development}
Masashi Misawa, Shin-ei Kudo, Yuichi Mori, Kinichi Hotta, Kazuo Ohtsuka, Takahisa Matsuda, Shoichi Saito, Toyoki Kudo, Toshiyuki Baba, Fumio Ishida, et~al.,
\newblock ``Development of a computer-aided detection system for colonoscopy and a publicly accessible large colonoscopy video database (with video),''
\newblock {\em Gastrointestinal endoscopy}, vol. 93, no. 4, pp. 960--967, 2021.

\bibitem{ji2022video}
Ge-Peng Ji, Guobao Xiao, Yu-Cheng Chou, Deng-Ping Fan, Kai Zhao, Geng Chen, and Luc Van~Gool,
\newblock ``Video polyp segmentation: A deep learning perspective,''
\newblock {\em Machine Intelligence Research}, vol. 19, no. 6, pp. 531--549, 2022.

\bibitem{vazquez2017benchmark}
David V{\'a}zquez, Jorge Bernal, F~Javier S{\'a}nchez, Gloria Fern{\'a}ndez-Esparrach, Antonio~M L{\'o}pez, Adriana Romero, Michal Drozdzal, Aaron Courville, et~al.,
\newblock ``A benchmark for endoluminal scene segmentation of colonoscopy images,''
\newblock {\em Journal of healthcare engineering}, vol. 2017, 2017.

\bibitem{zhao2023unipc}
Wenliang Zhao, Lujia Bai, Yongming Rao, Jie Zhou, and Jiwen Lu,
\newblock ``Unipc: A unified predictor-corrector framework for fast sampling of diffusion models,''
\newblock {\em arXiv preprint arXiv:2302.04867}, 2023.

\end{thebibliography}

\end{document}